\documentclass[runningheads]{llncs}
\usepackage[T1]{fontenc}
\usepackage{graphicx,verbatim}
\usepackage{amssymb}
\usepackage{amsmath}
\usepackage{upgreek}
\usepackage{booktabs}
\usepackage{multirow}
\usepackage{array}
\usepackage{graphicx}
\usepackage{float}
\usepackage{colortbl}
\usepackage{hyperref}
\definecolor{highlightblue}{RGB}{219, 234, 254}

\begin{document}
\title{Learning to \texttt{Unify} Deformable Shape and Texture Representations for Cardiac Video Classification}

\author{Tonmoy Hossain\inst{1} \and
Miaomiao Zhang\inst{1,2}}

\authorrunning{Hossain and Zhang}

\institute{Department of Computer Science, University of Virginia, Virginia, USA \and
Department of Electrical and Computer Engineering, University of Virginia, USA
}
   
\maketitle              
\begin{abstract}
Deformable shape representations have proven to be robust complements to texture features in cardiac image classification, offering geometric priors that are invariant to imaging artifacts and intensity variations. However, existing deep networks perform simple concatenation to combine these distinct feature representations, which neither fully exploits their complementary nature nor learns cross-modal feature dependencies. Furthermore, this results in uniform attention across all timepoints; hence ignoring the varying diagnostic importance across the cardiac phases. In this paper, we propose a novel cardiac video classification model that, for the first time, \textit{learns temporal features in an integrated space of deformable shape and image texture representations}. In particular, we design a bi-directional cross-attention in the latent space to fuse latent deformable shape and image features, allowing each modality to adaptively weight the other based on spatio-temporal correspondence. In contrast to current methods that apply uniform weighting across all the cardiac phases, our approach learns to dynamically adjust the contributions of shape and texture representations, derived from images, over time. We demonstrate state-of-the-art classification performance on a cine cardiac magnetic resonance (CMR) video dataset, achieving improved interpretability from attention mechanisms that identify diagnostically critical cardiac phases and modality contributions. Our code is publicly available at \href{https://github.com/tonmoy-hossain/ShapeFuse}{\tt https://github.com/tonmoy-hossain/ShapeFuse}.


\end{abstract}

\section{Introduction}

Cardiac MRI enables non-invasive assessment of myocardial function and is widely used for disease diagnosis, yet automated classification from CMR videos remains challenging~\cite{cau2024cine,jacob2025deep,jayakumar2023sadir,xie2024automatic}. Recent models capture temporal dependencies across frames but solely operate on raw image intensities. As a result, they mostly focus on texture-based features without understanding the underlying myocardial deformations~\cite{amyar2023gadolinium,clough2019global,zheng2019explainable}. However, pathological conditions, such as cardiomyopathy and myocarditis manifest as regional wall motion abnormalities, asymmetric contractility, and altered strain patterns~\cite{foley2018quantitative,gao2023cardiac}. Therefore, geometric shape features captured in motion or deformation fields often encode richer information that can complement intensity-based representations alone~\cite{qin2023generative,xing2024lamod,xing2024multimodal}. 

Recent works have explored integrating deformable shape representations with intensity features for neurodegenerative disease classification, where geometric priors have proven to be robust complements to texture-based representations~\cite{hossain2024invariant,hossain2025corld,wang2022geo}. Extending this paradigm to cardiac video analysis, however, is not straightforward given the temporal nature of the cardiac cycle. A critical challenge lies in how to combine these two intertwined representations, as simple operations of element-wise addition or concatenation can ignore cross-modal feature dependencies. In particular, such approaches do not model how shape and texture jointly interact across the cardiac cycle, which limits their ability to learn complementary physiological information.

For instance, deformation features are most informative during dynamic transition phases of the cardiac cycle, particularly from end-diastole through peak systolic contraction and early relaxation, where myocardial motion changes are most evident. In contrast, texture-based intensity patterns are less sensitive to these functional dynamics and may become unreliable under conditions such as low contrast-to-noise ratio or motion-induced imaging artifacts~\cite{alis2021influence}. Despite the clear clinical relevance, prior works have not explicitly learned to model the dynamic interaction between myocardial deformation and image texture for disease classification.

Building on the premise of integrating deformable shape and texture representations, this paper presents {\tt ShapeFuse}, the first framework to jointly learn the fusion of deformable shape and texture representations in a shared latent space for cardiac video classification. In contrast to existing methods that rely on static concatenation or addition, {\tt ShapeFuse} explicitly learns cross-modal dependencies and dynamically weights modality contributions across the cardiac cycle. To achieve this, we propose a three-fold contribution:
\begin{itemize}
    \item Develop a bidirectional cross-modal temporal attention mechanism that learns mutual dependencies between latent deformable shape and texture representations across the cardiac cycle.

    \item Design an adaptive fusion gate that learns to weight shape and texture contributions at each cardiac phase, coupled with a learned temporal pooling that prioritizes diagnostically relevant timepoints for classification.

    \item Demonstrate SOTA cardiac video classification on real-world cine CMR videos, with interpretable attention maps that localize diagnostically critical cardiac phases and quantify modality-wise contributions. 
\end{itemize}

\section{Background: Deformation Learning from Image Pairs}

Deformation learning from image pairs aims to establish spatial correspondence between a source image $\mathcal{S}$ and a target image $\mathcal{T}$ by estimating a transformation that aligns $\mathcal{S}$ to $\mathcal{T}$. By following the principles of diffeomorphic registration, the transformation $\phi$ can be derived by mapping from $\mathcal{S}$ to $\mathcal{T}$ through integrating a stationary velocity field (SVF~\cite{arsigny2006log}) $v$ via the flow equation
\begin{equation}
\frac{d\phi(t)}{dt} = v \circ \phi(t), \quad \phi(0) = \text{Id},
\label{eq:svf}
\end{equation}
where $v$ remains constant over time, and Id denotes an identity transformation. The solution at $t = 1$ is denoted by $\phi = \exp(v)$, computed via scaling-and-squaring as $\phi \approx (\text{Id} + v/2^K)^{2^K}$. While we adopt the SVF formulation in this work, our proposed framework generalizes to other diffeomorphic parameterizations such as time-varying velocity fields. Following this transformation to deform $\mathcal{S}$ onto $\mathcal{T}$, we minimize the energy function
\begin{equation}
E = \frac{1}{\sigma^2} \mathcal{D}[\mathcal{S} \circ \phi(v), \mathcal{T}] + \|\nabla v\|_1, \quad \text{s.t. Eq.~\eqref{eq:svf}},
\label{eq:energy}
\end{equation}
where $\mathcal{D}[\cdot, \cdot]$ measures image dissimilarity between the warped source $\mathcal{S} \circ \phi(v)$ and target $\mathcal{T}$, $\sigma^2$ represents noise variance, and the regularization term $\|\nabla v\|_1$ enforces smoothness of the velocity field. For the dissimilarity metric $\mathcal{D}$, we can employ sum of squared differences (SSD)~\cite{beg2005computing}, normalized cross-correlation (NCC)~\cite{avants2008symmetric}, or mutual information (MI)~\cite{wells1996multi}.

\section{Our Method: {\tt ShapeFuse}}
\begin{figure}[!t]
    \centering
    \includegraphics[width=1.0\linewidth, trim=0 0 0 0]{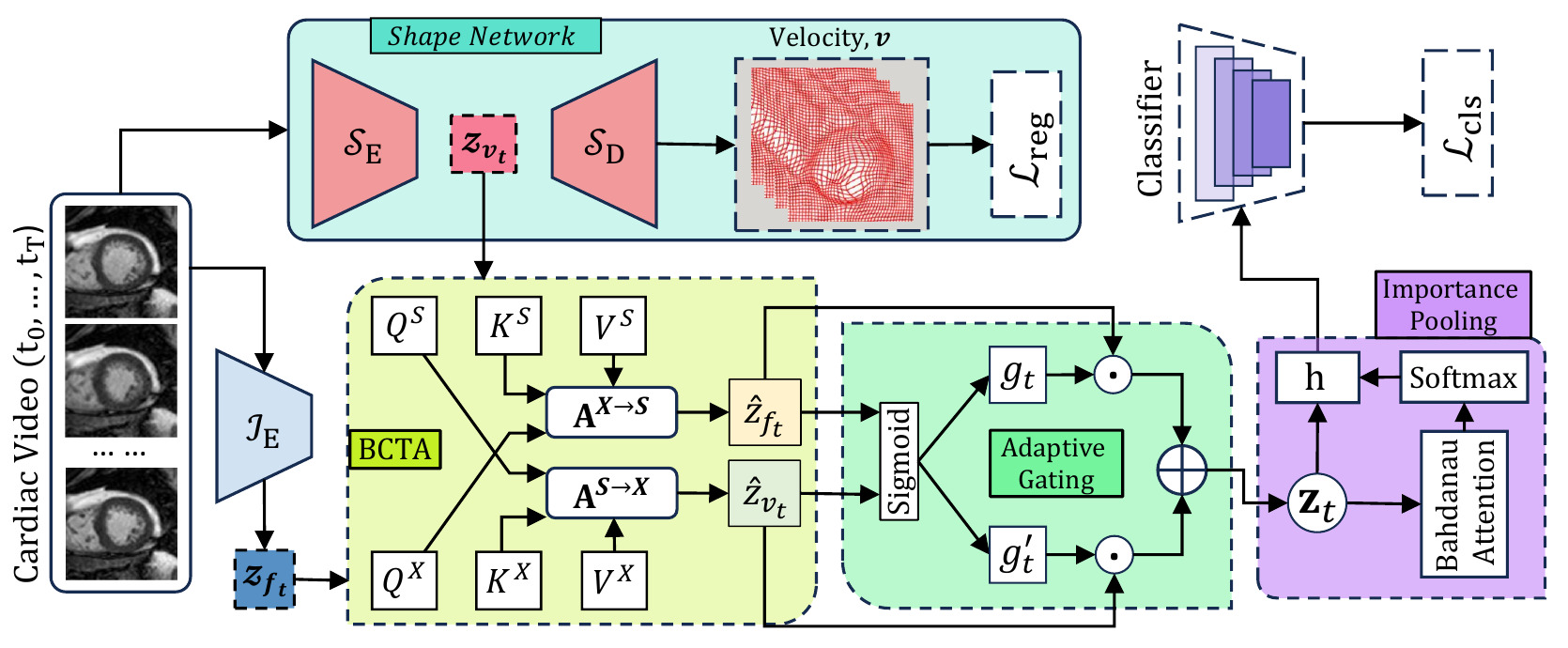}
    \caption{An overview of our cardiac video classification model.}
    \label{fig:model}
\end{figure}

This section introduces ShapeFuse, a novel framework that unifies representations of deformable shape and image intensity for improved cardiac video classification. At the core of ShapeFuse is a {\em latent shape-aware feature fusion} module that consists of two key components: (i) a {\em bidirectional cross-modal temporal attention} mechanism to enable dynamic interactions between shape and image features over time, and (ii) an {\em adaptive gating and diagnostic importance pooling} strategy to selectively emphasize the most clinically informative features and temporal phases for classification. An overview of the proposed architecture is illustrated in Fig.~\ref{fig:model}.

\noindent{\textbf{Problem Definition.}} Given a CMR dataset, $\mathcal{I} = \{I_0, I_1, \ldots, I_T\}$ with $T+1$ frames, our objective is to predict a disease label $y \in \mathcal{Y}$. To leverage geometric priors that complement texture-based representations, we estimate frame-wise velocity fields $\mathbf{S}$ from an encoder-decoder network $(\theta_S^E, \theta_S^D)$, and derive texture representations $\mathbf{X}$ from an image encoder $\theta_I^E$. \textit{The core challenge is to unify these two distinct yet intertwined feature representations in the latent space such that their cross-modal dependencies and temporal relationships across cardiac phases are learned jointly, rather than treating them as independent modalities.}

\subsection{Latent Shape-aware Feature Fusion} Given the input sequence $\mathcal{I}$, we estimate the latent geometric deformation of a target frame $I_t$ at each timepoint $t$ relative to a source frame $I_0$ as $\mathbf{z}_{v_t} \in \mathbb{R}^{C \times H \times W}$, inferred from the shape encoder $\theta_S^E$. In parallel, we encode each frame $I_t$ through $\theta_I^E$ to obtain its latent texture representation $\mathbf{z}_{f_t}$ in a shared latent space, enabling cross-modal interaction with the latent shape features.\\

\noindent\textbf{Bidirectional Cross-Modal Temporal Attention.} While SOTA models extensively adopt element-wise summation or concatenation as the fusion strategy, neither exploits the complementary nature of the two modalities nor learns cross-modal feature dependencies. As a result, these approaches impose uniform weighting across all timepoints, ignoring the varying diagnostic importance across cardiac phases. Each latent map is first flattened and linearly projected to a token of dimension $d$, so that $\mathbf{z}_{v_t}, \mathbf{z}_{f_t} \in \mathbb{R}^{d}$. To this end, we formulate a bidirectional cross-modal temporal attention over the shape sequence $\mathbf{S} = [\mathbf{z}_{v_0}, \ldots, \mathbf{z}_{v_T}]$ and texture sequence $\mathbf{X} = [\mathbf{z}_{f_0}, \ldots, \mathbf{z}_{f_T}]$, where each modality has its own learnable query, key, and value projections, denoted as $\{\mathbf{W}^{S}_{Q}, \mathbf{W}^{S}_{K}, \mathbf{W}^{S}_{V}\}$ and $\{\mathbf{W}^{X}_{Q}, \mathbf{W}^{X}_{K}, \mathbf{W}^{X}_{V}\} \in \mathbb{R}^{d \times d}$, with $d$ denoting the latent dimension. This allows each modality to attend to the other across all timepoints $i, j \in \{0, 1, \dots, T\}$ by
\begin{equation}
\begin{aligned}
    Q_i^S &= \mathbf{W}_Q^S\,\mathbf{z}_{v_i}, \quad K_j^S = \mathbf{W}_K^S\,\mathbf{z}_{v_j}, \quad V_j^S = \mathbf{W}_V^S\,\mathbf{z}_{v_j}, \\
    Q_i^X &= \mathbf{W}_Q^X\,\mathbf{z}_{f_i}, \quad K_j^X = \mathbf{W}_K^X\,\mathbf{z}_{f_j}, \quad V_j^X = \mathbf{W}_V^X\,\mathbf{z}_{f_j}.
\end{aligned}
\end{equation}
The resulting attention maps $\mathbf{A}^{S \to X}, \mathbf{A}^{X \to S} \in \mathbb{R}^{(T+1) \times (T+1)}$ explicitly parameterize cross-phase dependencies, where $i$ and $j$ index the query and key timepoints, each entry $\mathbf{A}^{S \to X}_{ij}$ captures the relevance of texture at timepoint $j$ to shape at timepoint $i$, and vice versa, as
\begin{equation}
    \mathbf{A}^{S \to X}_{ij} = \frac{\exp\left(Q_i^S \cdot K_j^X / \sqrt{d}\right)}{\sum_{j'} \exp\left(Q_i^S \cdot K_{j'}^X / \sqrt{d}\right)}, \quad
    \mathbf{A}^{X \to S}_{ij} = \frac{\exp\left(Q_i^X \cdot K_j^S / \sqrt{d}\right)}{\sum_{j'} \exp\left(Q_i^X \cdot K_{j'}^S / \sqrt{d}\right)}.
\end{equation}
We then obtain the attended representations $\tilde{\mathbf{z}}_{v_t}$ and $\tilde{\mathbf{z}}_{f_t}$ as
\begin{equation}
    \tilde{\mathbf{z}}_{v_t} = \sum_{j=0}^{T} \mathbf{A}^{S \to X}_{tj}\,V_j^X, \quad \tilde{\mathbf{z}}_{f_t} = \sum_{j=0}^{T} \mathbf{A}^{X \to S}_{tj}\,V_j^S. 
\end{equation}
Similar to~\cite{vaswani2017attention}, the final feature representations are stabilized using residual connections and layer normalization modules, i.e., $\hat{\mathbf{z}}_{v_t} = \text{LN}(\mathbf{z}_{v_t} + \tilde{\mathbf{z}}_{v_t})$ and $\hat{\mathbf{z}}_{f_t} = \text{LN}(\mathbf{z}_{f_t} + \tilde{\mathbf{z}}_{f_t})$.

\noindent\textbf{Adaptive Gating and Diagnostic Importance Pooling.} Since the diagnostic relevance of shape and texture is inherently timepoint-dependent, symmetric combinations of $\hat{\mathbf{z}}_{v_t}$ and $\hat{\mathbf{z}}_{f_t}$ would ignore this critical property. We therefore introduce a per-timepoint adaptive gate $g_t$ that dynamically controls the relative contribution of each modality as
\begin{equation}
    g_t = \sigma\!\left(\mathbf{W}_g \left[\hat{\mathbf{z}}_{v_t} \,\|\, \hat{\mathbf{z}}_{f_t}\right] + b_g\right), \quad \mathbf{z}_t = g_t \odot \hat{\mathbf{z}}_{v_t} + (1 - g_t) \odot \hat{\mathbf{z}}_{f_t},
\end{equation}
where $\sigma(\cdot)$ is the sigmoid function, $\mathbf{W}_g \in \mathbb{R}^{d \times 2d}$ and $\mathbf{b}_g \in \mathbb{R}^{d}$ are the learnable gating weight and bias, $\|$ denotes concatenation, $\odot$ is element-wise multiplication, $g_t' = 1 - g_t$ is the complementary gate applied to the latent texture representation $\hat{\mathbf{z}}_{f_t}$. In contrast to previous uniform temporal pooling that treats all cardiac phases as equally informative, we further aggregate the fused sequence $\{\mathbf{z}_0, \ldots, \mathbf{z}_T\}$ via a learned diagnostic importance weighting by following Bahdanau Attention~\cite{bahdanau2014neural}
\begin{equation}
    \alpha_t = \frac{\exp\left(\mathbf{w}^\top \tanh(\mathbf{W}_h\,\mathbf{z}_t)\right)}{\sum_{t'} \exp\left(\mathbf{w}^\top \tanh(\mathbf{W}_h\,\mathbf{z}_{t'})\right)}, \quad \mathbf{h} = \sum_{t=0}^{T} \alpha_t\,\mathbf{z}_t,
\end{equation}
where $\mathbf{w} \in \mathbb{R}^d$ and $\mathbf{W}_h \in \mathbb{R}^{d \times d}$ are learnable parameters, $t'$ indexes the timepoints $\{0, \ldots, T\}$ over which the attention weights are normalized, and $\alpha_t$ reflects the learned diagnostic importance of each cardiac phase.

\subsection{Training Objective}

We train the classification framework in two stages. Following Eqs.~\eqref{eq:svf}-~\eqref{eq:energy}, we first optimize the shape network $(\theta_S^E, \theta_S^D)$ over the velocity fields $v$ by minimizing the objective function defined as 

\begin{equation}
\mathcal{L}_{\text{reg}}(\theta_S^E, \theta_S^D) = \sum_{t=1}^{T} \left[
\frac{1}{\sigma^2} \mathcal{D}[I_0 \circ \phi_t(\mathbf{z}_{v_t}(\theta_S^E, \theta_S^D)), I_t]
+ \lVert \nabla v_t \rVert_1 \right] + \text{reg}(\theta_S^E, \theta_S^D),
\label{eq:registration}
\end{equation}
where $\mathcal{D}[\cdot,\cdot]$ denotes the sum-of-squared differences between the warped source $I_0\circ\phi_t$ and target $I_t$, $\sigma^{2}$ denotes noise variance, $\|\nabla v_t\|_1$ enforces smoothness of the velocity field, and \text{reg(·)} denotes network parameter regularization.

During the second stage, the latent representations $\{\mathbf{z}_{v_t}\}$ learned from Eq.~\eqref{eq:registration} are held fixed. The image encoder $\theta_I^E$, cross-modal attention $\{\mathbf{W}_Q, \mathbf{W}_K, \mathbf{W}_V\}$, adaptive gating $\{\mathbf{W}_g, \mathbf{b}_g\}$, diagnostic importance pooling $\{\mathbf{w}, \mathbf{W}_h\}$, and classifier $\theta_c$, collectively denoted as $\Theta$, are jointly optimized via a binary cross-entropy loss over the aggregated representation $\mathbf{h}$ as
\begin{equation}
\mathcal{L}_{\text{cls}}(\Theta) = -\frac{1}{N}\sum_{n=1}^{N} \left[ y_n \log \hat{y}_n + (1 - y_n) \log (1 - \hat{y}_n) \right] + \text{reg}(\Theta),
\end{equation}
where $\hat{y}_n = \theta_c(\mathbf{h}_n)$ is the predicted probability for sample $n$ and $\text{reg}(\Theta)$ denotes a regularization term on the parameters $\Theta$. 

\section{Experimental Evaluation}
We evaluate {\tt ShapeFuse} on a cine CMR video dataset~\cite{wang2022ai}. This dataset consists of $510$ cine CMR video sequences from $125$ subjects, each accompanied by manually delineated left ventricular myocardium segmentation maps. Nearly $40\%$ of subjects present with myocardial scar-induced wall motion abnormalities, providing a clinically relevant class imbalance. All sequences span a complete cardiac cycle of $24$ time frames and are standardized to $224 \times 224 \times 24$.

\subsection{Experiments}
\noindent{\textbf{Effectiveness of {\tt ShapeFuse} in Image Classification.}} We validate our model on classification tasks, adopting convolutional architectures (ResNet~\cite{he2016deep}, EfficientNet~\cite{tan2019efficientnet}, and  DenseNet~\cite{huang2017densely}) and transformer-based model (ViT~\cite{dosovitskiy2020image}) as image encoder backbones. Performance is evaluated using micro-averaged accuracy and F1-score. We ablate the fusion module by replacing {\tt ShapeFuse} with addition, concatenation~\cite{wang2022geo}, weighted~\cite{chen2019synergistic}, bilinear~\cite{ni2022space}, and temporal attention-based fusion strategies, evaluated across all encoder backbones and two registration networks, including VoxelMorph (VM)~\cite{balakrishnan2019voxelmorph} and TLRN~\cite{wu2024tlrn}.

\noindent{\textbf{Interpretability Analysis.}} We visualize Grad-CAM activation maps across cardiac timepoints and compare them with those produced by models employing conventional fusion strategies to assess how different fusion mechanisms influence spatial attention and feature localization.

\noindent{\textbf{Cross-Modal Attention and Gating Analysis.}} We further analyze the learned cross-modal attention maps and adaptive gating distributions to examine the dynamic interaction between shape and texture modalities across the cardiac phases.\\

\noindent{\textit{Implementation Details.}} Both shape and texture features are projected into a shared hidden space of dimension $d=512$, fused via bidirectional multi-head cross-attention ($8$ heads) with residual connections and layer normalization. We pass the fused representation through a three-layer fully connected classifier with hidden dimensions $512$, $256$, and $64$, with batch normalization, LeakyReLU activations, and dropout ($p=0.3$). We kept the shape encoder frozen during classifier training, but joint training can also be done. We optimize the model using AdamW~\cite{kingma2014adam} with a learning rate of $1\times10^{-5}$ and weight decay of $0.01$, with a ReduceLROnPlateau scheduler (factor $0.5$). We train the model up to $500$ epochs with early stopping (patience $20$, $\delta=0.001$) based on validation loss. We utilize a 40GB NVIDIA A100 Tensor Core GPU for training and testing.

\subsection{Results}
Tab.~\ref{tab:clf_perf} reports the classification performance of {\tt ShapeFuse} against established fusion strategies across multiple image encoder backbones and two registration networks. Under both VM and TLRN registration backbones, {\tt ShapeFuse} consistently outperforms all competing fusion strategies, demonstrating that explicitly modeling cross-modal dependencies yields meaningful gains over naive approaches such as concatenation and element-wise addition, which fail to reliably improve over the shape-only baseline. The consistent gains across all image encoders and registration backbones confirm that {\tt ShapeFuse} is an effective and generalizable cross-modal fusion module for cardiac video classification.

\begin{table}[t]
\centering
\caption{Classification performance comparison of {\tt ShapeFuse} against competing fusion strategies across multiple image encoder backbones and registration networks.}
\begin{tabular}{c lcc c cc c cc c cc}
\toprule
\multicolumn{2}{c}{Backbone}
& \multicolumn{2}{c}{\textit{ResNet}}
& & \multicolumn{2}{c}{\textit{EfficientNet}}
& & \multicolumn{2}{c}{\textit{DenseNet}}
& & \multicolumn{2}{c}{\textit{ViT}} \\
\cmidrule(lr){1-2} \cmidrule(lr){3-4} \cmidrule(lr){6-7} \cmidrule(lr){9-10} \cmidrule(lr){12-13}
\multicolumn{2}{c}{Metrics}
& Acc & F1-sc
& & Acc & F1-sc
& & Acc & F1-sc
& & Acc & F1-sc \\
\midrule
& Image
& $0.764$ & $0.751$ & \vline
& $0.787$ & $0.822$ & \vline
& $0.797$ & $0.802$ & \vline
& $0.786$ & $0.779$ \\
\midrule
\multirow{7}{*}{\rotatebox{90}{\textit{VM}}}
& Shape
& $0.809$ & $0.825$ & \vline
& $0.809$ & $0.825$ & \vline
& $0.809$ & $0.825$ & \vline
& $0.809$ & $0.825$ \\
& + Add
& $0.807$ & $0.884$ & \vline
& $0.787$ & $0.822$ & \vline
& $0.820$ & $0.849$ & \vline
& $0.819$ & $0.808$ \\
& + Concat~\cite{wang2022geo}
& $0.764$ & $0.800$ & \vline
& $0.809$ & $0.841$ & \vline
& $0.775$ & $0.818$ & \vline
& $0.830$ & $0.824$ \\
& + Weighted~\cite{chen2019synergistic}
& $0.832$ & $0.857$ & \vline
& $0.787$ & $0.816$ & \vline
& $0.753$ & $0.804$ & \vline
& $0.852$ & $0.827$ \\
& + Bilinear~\cite{ni2022space}
& $0.832$ & $0.857$ & \vline
& $0.820$ & $0.846$ & \vline
& $0.787$ & $0.829$ & \vline
& $0.852$ & $0.834$ \\
& + Attention
& $0.787$ & $0.808$ & \vline
& $0.820$ & $0.852$ & \vline
& $0.843$ & $0.865$ & \vline
& $0.830$ & $0.844$ \\
\rowcolor{highlightblue}
& + \textbf{{\tt ShapeFuse}}
& $\mathbf{0.888}$ & $\mathbf{0.902}$ & \vline
& $\mathbf{0.843}$ & $\mathbf{0.857}$ & \vline
& $\mathbf{0.854}$ & $\mathbf{0.871}$ & \vline
& $\mathbf{0.869}$ & $\mathbf{0.858}$ \\
\midrule
\multirow{7}{*}{\rotatebox{90}{\textit{TLRN}}}
& Shape
& $0.797$ & $0.836$ & \vline
& $0.797$ & $0.836$ & \vline
& $0.797$ & $0.836$ & \vline
& $0.797$ & $0.836$ \\
& + Add
& $0.854$ & $0.871$ & \vline
& $0.831$ & $0.857$ & \vline
& $0.820$ & $0.852$ & \vline
& $0.831$ & $0.848$ \\
& + Concat~\cite{wang2022geo}
& $0.820$ & $0.840$ & \vline
& $0.843$ & $0.860$ & \vline
& $0.831$ & $0.860$ & \vline
& $0.807$ & $0.790$ \\
& + Weighted~\cite{chen2019synergistic}
& $0.854$ & $0.876$ & \vline
& $0.865$ & $0.882$ & \vline
& $0.820$ & $0.852$ & \vline
& $0.852$ & $0.848$ \\
& + Bilinear~\cite{ni2022space}
& $0.876$ & $0.884$ & \vline
& $0.854$ & $0.874$ & \vline
& $0.855$ & $0.871$ & \vline
& $0.841$ & $0.837$ \\
& + Attention
& $0.842$ & $0.857$ & \vline
& $0.842$ & $0.844$ & \vline
& $0.854$ & $0.846$ & \vline
& $0.831$ & $0.820$ \\
\rowcolor{highlightblue}
& + \textbf{{\tt ShapeFuse}}
& $\mathbf{0.899}$ & $\mathbf{0.901}$ & \vline
& $\mathbf{0.876}$ & $\mathbf{0.884}$ & \vline
& $\mathbf{0.876}$ & $\mathbf{0.891}$ & \vline
& $\mathbf{0.862}$ & $\mathbf{0.850}$ \\

\bottomrule
\end{tabular}
\label{tab:clf_perf}
\end{table}

\begin{figure}[htbp]
    \centering
    \includegraphics[width=\linewidth, trim=0 0 0 0]{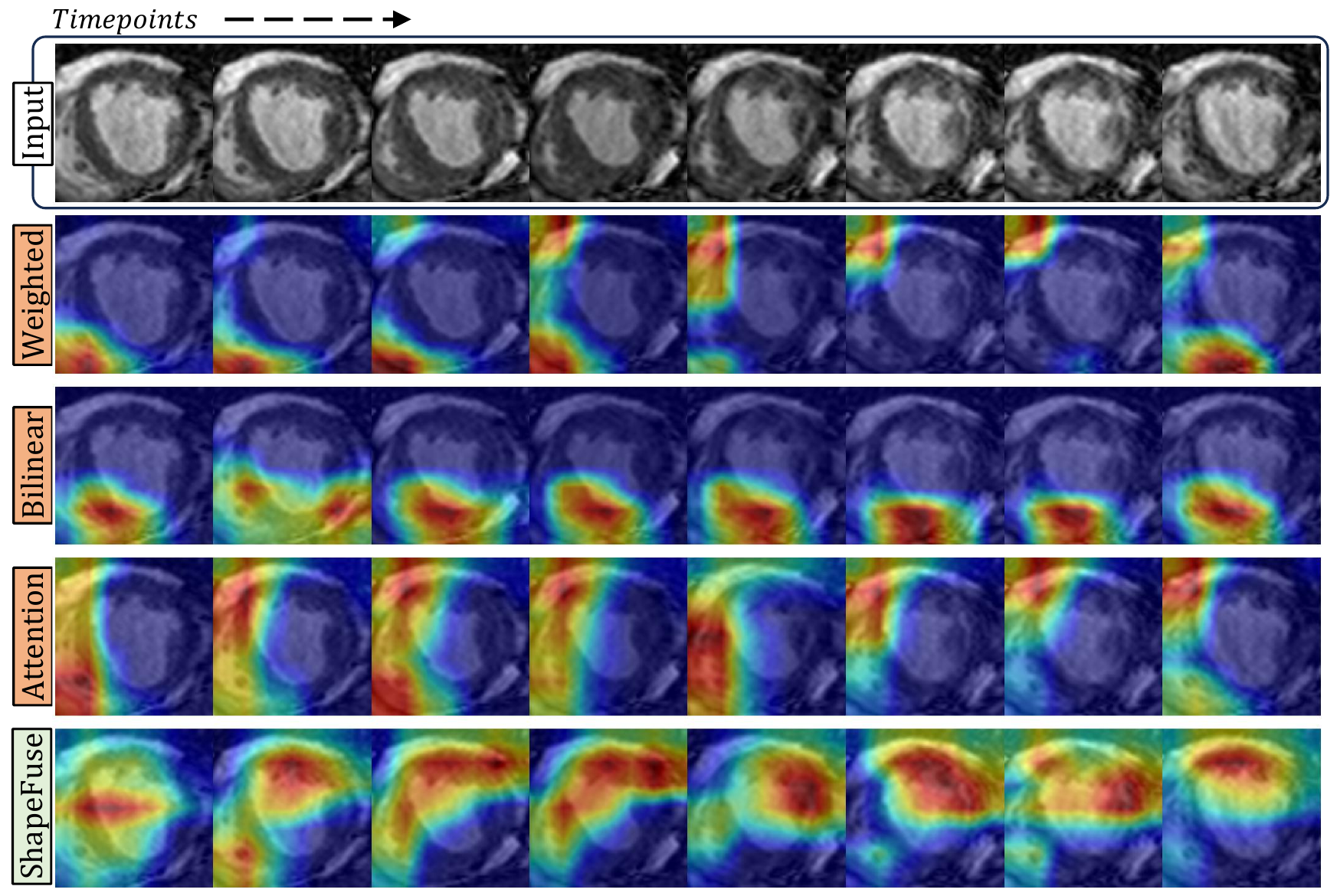}
    \caption{Grad-CAM activation maps across the cardiac timepoints for {\tt ShapeFuse} and baseline fusion strategies.}
    \label{fig:gradcam_vizs}
\end{figure}

Fig.~\ref{fig:gradcam_vizs} visualizes Grad-CAM activation maps across cardiac timepoints for different fusion strategies. Unlike competing methods that produce diffuse or spatially inconsistent activations, {\tt ShapeFuse} consistently highlights the myocardial region across all timepoints, which is particularly meaningful given that myocardial motion throughout the cardiac cycle is an established indicator of scar presence and regional wall motion abnormalities. This targeted localization demonstrates that {\tt ShapeFuse} captures temporally evolving myocardial deformation patterns most relevant to scar diagnosis, yielding both superior classification performance and more clinically interpretable spatial attention maps.

\begin{figure}[htbp]
    \centering
    \includegraphics[width=\linewidth, trim=0 0 0 0]{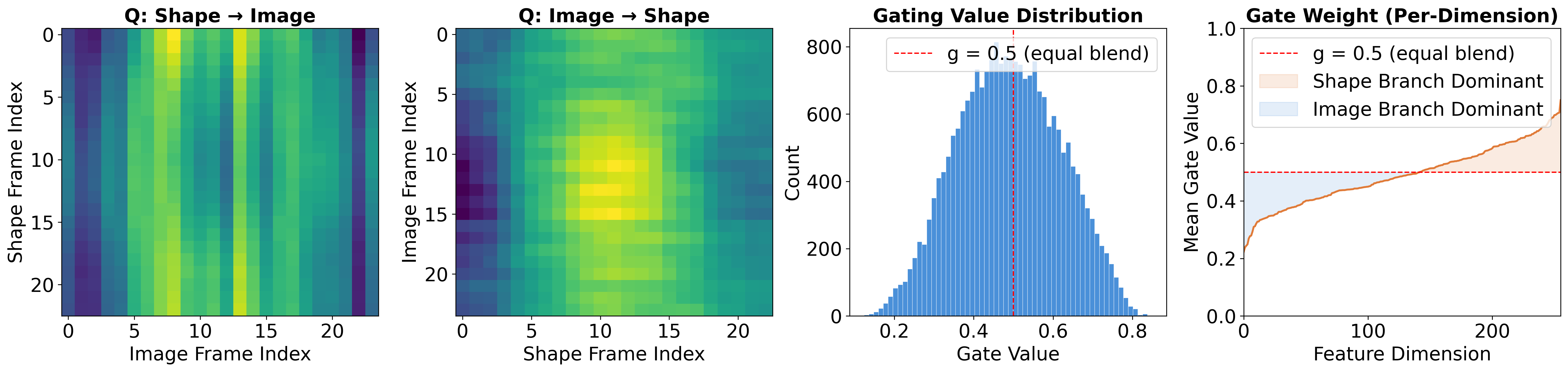}
    \caption{Analysis of the learned cross-modal attention and adaptive gating strategy.}
    \label{fig:attention}
\end{figure}

Fig.~\ref{fig:attention} validates {\tt ShapeFuse}'s learned cross-modal attention and adaptive gating. The shape-to-image attention shows deformation features consistently attending to mid-systolic texture frames, suggesting peak contraction carries the most discriminative evidence for scar classification. The image-to-shape attention follows a diagonal pattern, indicating texture representations seek phase-aligned geometric context. This asymmetry confirms that BCTA captures complementary cross-modal interactions that symmetric fusion operators cannot model. The gating distribution centered slightly below $g = 0.5$ reflects that texture representations benefit more from geometric grounding, consistent with the clinical challenge of identifying scar-induced hypokinesia from intensity alone. The per-dimension gate weight profile further shows clean separation between shape/image-dominant dimensions, confirming the adaptive gate allocates the two modalities into complementary subspaces, providing evidence that {\tt ShapeFuse}'s learned fusion directly translates into the classification gains observed in Tab.~\ref{tab:clf_perf}.

\section{Conclusion}
This paper presents {\tt ShapeFuse}, a novel framework for cardiac video classification that jointly learns the fusion of deformable shape and texture representations in a shared latent space. By replacing naive concatenation with bidirectional cross-modal temporal attention and adaptive gating, {\tt ShapeFuse} explicitly models cross-modal dependencies and dynamically weights modality contributions across the cardiac cycle, achieving state-of-the-art classification performance on a cine CMR dataset with improved interpretability. 

In future work, we plan to extend {\tt ShapeFuse} to multi-label pathology classification, where multiple co-occurring conditions require simultaneous localization, and to explore self-supervised pretraining of the cross-modal attention module to reduce reliance on labeled data in low-resource clinical settings.\\

\noindent \textbf{Acknowledgments.} This work was supported by NSF CAREER Grant $2239977$ and NIH 1R21EB032597.\\

\noindent \textbf{Disclosure of Interests.} The authors have no competing interests to declare that
are relevant to the content of this article.

\bibliographystyle{splncs04}
\bibliography{Paper-6046}
\end{document}